\documentclass[sigconf]{acmart}
\AtBeginDocument{%
  \providecommand\BibTeX{{%
    \normalfont B\kern-0.5em{\scshape i\kern-0.25em b}\kern-0.8em\TeX}}}

\setcopyright{acmcopyright}
\copyrightyear{2023}
\acmYear{2023}
\acmDOI{XXXXXXX.XXXXXXX}

\acmConference[KDD]{Foundations and Applications in Large-scale AI Models
-Pre-training, Fine-tuning, and Prompt-based Learning}{August,
  2023}{Long Beach, California}
\acmPrice{15.00}
\acmISBN{978-1-4503-XXXX-X/18/06}

\usepackage{hyperref}

\begin{document}

\title{Text-to-Video: a Two-stage Framework for Zero-shot Identity-agnostic Talking-head Generation}


\author{Zhichao Wang}
\affiliation{%
  \institution{Salesforce}
  \country{}}
\email{zhichaowang@salesforce.com}

\author{Mengyu Dai}
\affiliation{%
  \institution{Salesforce}
  \country{}}
\email{mdai@salesforce.com}

\author{Keld Lundgaard}
\affiliation{%
  \institution{Salesforce}
  \country{}}
\email{klundgaard@salesforce.com}

\renewcommand{\shortauthors}{Wang, Dai and Lundgaard}

\begin{abstract}
The advent of ChatGPT has introduced innovative methods for information gathering and analysis. However, the information provided by ChatGPT is limited to text, and the visualization of this information remains constrained. Previous research has explored zero-shot text-to-video (TTV) approaches to transform text into videos. However, these methods lacked control over the identity of the generated audio, i.e., not identity-agnostic, hindering their effectiveness. To address this limitation, we propose a novel two-stage framework for person-agnostic video cloning, specifically focusing on TTV generation. In the first stage, we leverage pretrained zero-shot models to achieve text-to-speech (TTS) conversion. In the second stage, an audio-driven talking head generation method is employed to produce compelling videos privided the audio generated in the first stage. This paper presents a comparative analysis of different TTS and audio-driven talking head generation methods, identifying the most promising approach for future research and development. Some audio and videos samples can be found in the following link: https://github.com/ZhichaoWang970201/Text-to-Video/tree/main.
\end{abstract}





\maketitle

\section{Introduction}
\label{sec:intro}
With the advent of ChatGPT, the field of search has undergone revolutionary transformations, enabling humans to acquire information in a more intuitive manner. However, the gathered information from ChatGPT is currently limited to text form, posing challenges for effective visualization. Consequently, there is a pressing need to develop techniques for TTV generation to address this issue and provide improved visualization tools for the acquired information. TTV involves generating videos based on given text, enabling enhanced visualization of the gathered information. By leveraging TTV techniques, the visualization of information becomes more accessible and facilitates a more comprehensive understanding of the content.

Previous researchers have conducted several researchs on TTV. These past researches have limited the specific identity of generated video with lots of training processes. However, gathering sufficient data for training such a neural network was tedious. As a result, the property of zeor-shot and identity-agnostic were desired. At the same time, researchers focused on zero-shot and identity-agonistic TTS and audio-driven talking head generation. However, there has not been a work on combining these two fields together for zero-shot identity-agonistic TTV. 

Previous researchers have dedicated efforts to the field of TTV, conducting several studies in this domain. However, these past research endeavors often encountered limitations when it came to generating videos specific to particular identities, requiring extensive training processes. Unfortunately, gathering sufficient data for training neural networks in this context proved to be a tedious task. Consequently, there emerged a need for TTV approaches that possess the desirable properties of zero-shot and identity-agnostic capabilities. Concurrently, researchers have also focused on zero-shot and identity-agnostic TTS and audio-driven talking head generation. Surprisingly, there has been a lack of efforts aimed at combining these two fields to achieve zero-shot identity-agnostic TTV. This research gap highlights the potential for novel approaches that can seamlessly integrate these fields and enable the generation of TTV content with zero-shot identity-agnostic capabilities.

This paper aims to conduct a comprehensive analysis and comparison between previous and current TTV methodologies. Firstly, we will delve into a comparative evaluation of earlier TTV techniques, examining their limitations and advancements in the field. Subsequently, we will summarize several state-of-the-art approaches in zero-shot identity-agnostic TTS and audio-driven talking head generation. These cutting-edge methods will be qualitatively assessed, considering their performance and effectiveness. Finally, we will outline potential future directions and research opportunities in the realm of zero-shot identity-agnostic TTV.  

\section{Related Work}
The initial part of the literature review would concentrate on TTS. A significant milestone in this domain was the development of Tacotron, which consisted of three key elements: an encoder, a decoder, and a vocoder \cite{wang2017tacotron}. The encoder extracted relevant information from the text and passed it to the decoder, which generated a mel spectrogram. Subsequently, the vocoder transformed the mel spectrogram into audible audio. Nonetheless, Tacotron had a limitation in that its decoder relied on an autoregressive model, resulting in slow inference speed. To overcome this limitation, Fastspeech introduced modifications to the decoder, transforming it into a non-autoregressive model and significantly accelerating the inference speed for TTS synthesis \cite{ren2019fastspeech}. These approaches initially generated mel-spectrograms and then transformed them into waveforms using a vocoder. However, in order to eliminate the need for this intermediate mel-spectrogram representation, a groundbreaking end-to-end neural network was introduced, which directly generates waveforms \cite{kim2021conditional}.

Nevertheless, the previous models could not specify the idenity of the generated audio and achieve zero-shot capabilities. To overcome these limitations, Jia et al. conducted a benchmark study on TTS. They proposed a transfer learning approach that utilized pretrained speaker encoders for speaker verification\cite{heigold2016end} and vocoders \cite{oord2016wavenet}. To train an end-to-end neural network, YourTTS was introduced to achieve zero-shot identity-agnostic TTS \cite{casanova2023yourTTS}. VALL-E presented a novel method for TTS by generating a neural codec for intermediate representation \cite{wang2023neural}. VALL-E was trained on a much larger dataset, and the performances of VALL-E overshadowed the previous methods on TTS.

The second part of the literature review delved into the realm of audio-driven talking head generation, which encompassed two distinct categories. The first category involved the utilization of head images for audio-driven talking head generation. For instance, Audio2Head extracted information from both the image and audio inputs to predict motion \cite{wang2021audio2head}. This generated motion prediction was subsequently employed for flow field prediction, serving as a foundation for video generation. Another approach focused on training talking head generation models on specific identities, incorporating a motion field transfer module to bridge the gap between the training identity and the inference identity \cite{wang2021oneshot}. StyleHEAT leveraged StyleGAN for high-resolution video generation in this context \cite{yin2022styleheat}. SadTalker represented head images using 3DMM coefficients and generated sequential 3DMM coefficients based on the audio input \cite{zhang2023sadtalker}. These sequential 3DMM coefficients were then applied for face rendering and video synthesis.

The second category of audio-driven talking head generation centered around video-based approaches. AD-Nerf utilized image frames from videos for training and modeled the upper body using two separate neural radiance fields defined by implicit functions \cite{guo2021adnerf}. VideoReTalking focused on achieving lip synchronization for modifying talking head videos \cite{cheng2022videoretalking}. Tang et al. accelerated the video generation process by decomposing the dynamic neural radiance field into three components: head, torso, and audio generation \cite{tang2022realtime}. Another study decomposed videos into shape, expression, and pose, replacing the expression in videos with the expression extracted from the audio input for video generation \cite{tang2022memories}.

The third part of the literature review centered on TTV approaches. ObamaNet, although effective, required extensive training time to generate arbitrary TTV spoken by Obama \cite{kumar2017obamanet}. To solve this problem, Write-a-Speaker reduced the training time for TTV by decomposing it into two stages: a speaker-independent stage and a speaker-specific stage \cite{li2021writeaspeaker}. Zhang et al. introduced Text2Video, a novel approach that directly generated audio from video \cite{zhang2022text2video}. Initially, audio was generated and aligned with the corresponding text. Using the generated text and audio, a pose dictionary was created to predict keypoint poses. Finally, the Vid2Vid GAN was applied for video generation. While all three works achieved impressive TTV results, they could not specify the identity of the speaker in either the audio or the video. To address this limitation, a new framework is proposed in this paper to tackle this problem effectively.

\section{TTV}

\begin{figure*}[tb!]
\centering
    \includegraphics[width=0.9\textwidth]{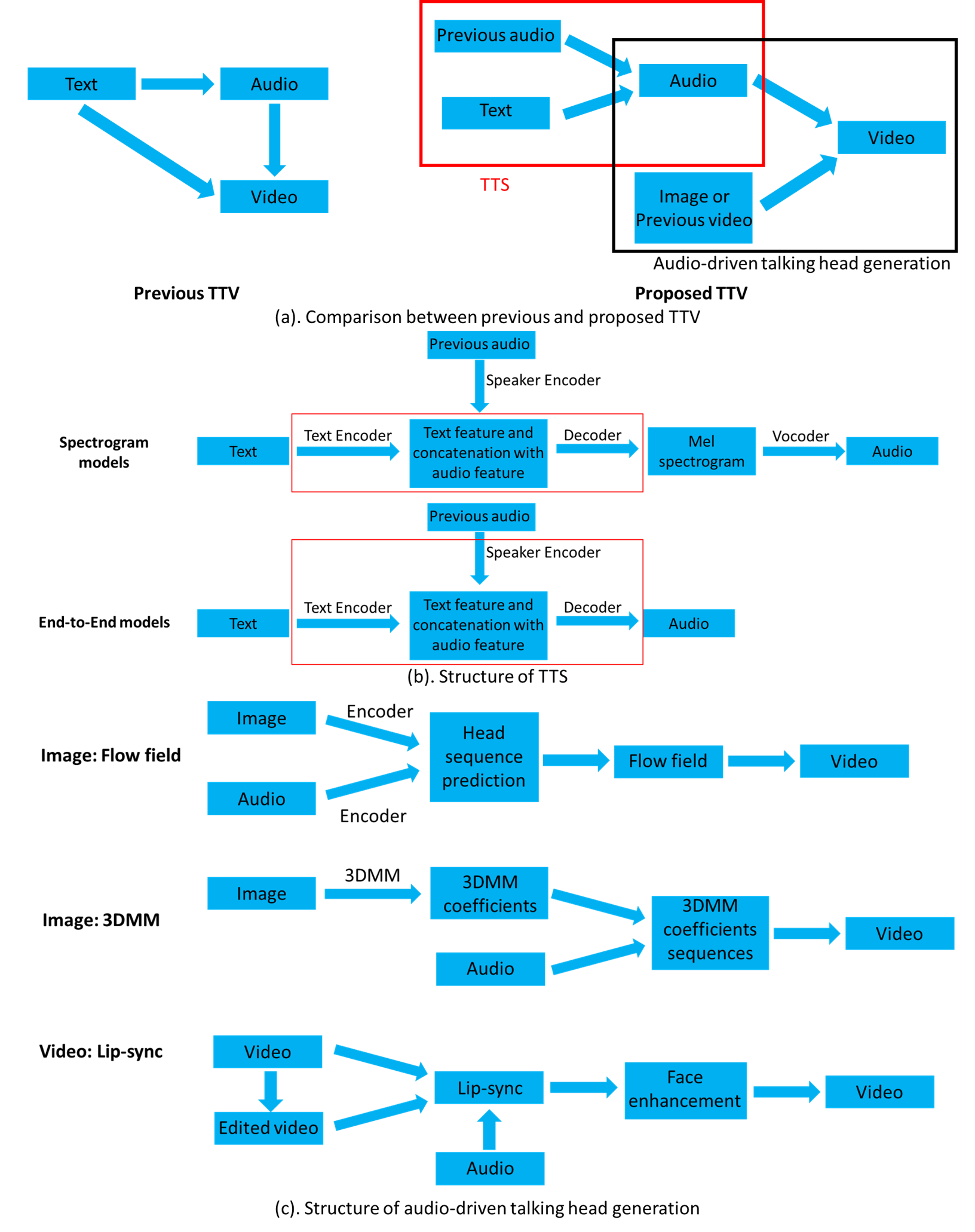}
    \caption{Comparsion of previous and current TTV and its components}
    \label{fig 1: Comparsion of previous and current TTV and its components}
\end{figure*}

This section will commence with a comprehensive comparison between the previous and current TTV approaches, depicted in part (a) of Figure \ref{fig 1: Comparsion of previous and current TTV and its components}. We will specifically review the Text2Video method proposed by Zhang et al. \cite{zhang2022text2video}. Their model followed a two-stage process, beginning with audio generation based on text input. This involved aligning the original text with the generated audio. Subsequently, the text combined with the generated audio were utilized to create a pose dictionary. By integrating the pose dictionary, audio, and vid2vidGAN, the TTV video was generated. While Text2Video has made notable advancements in the field, it is limited to generating audio and video specific to a predetermined identity utilized in the training process.

To address this limitation, this paper proposes a novel two-stage framework for TTV. The first stage leverages a well-established zero-shot identity-agnostic TTS framework. It utilizes the audio of a specific identity along with the text to generate speech. By mimicking the style of the provided audio, a new audio is generated with the specified identity speaking the given text. In the second stage, the generated audio is combined with a pre-existing image or video, integrating it into a well-developed audio-driven talking head generation system. Consequently, videos are generated that feature the specified identity speaking the provided text, without the need for additional training processes. Further details regarding the TTS and audio-driven talking head generation framework will be discussed in subsequent paragraphs.

There are two main directions in TTS research. In the traditional models depicted in the top region of part (b) of Figure \ref{fig 1: Comparsion of previous and current TTV and its components}, the neural network is consisted of an encoder, a decoder, and a vocoder. The encoder module encompasses a speech encoder, responsible for extracting information from previous audio, and a text encoder, which extracts information from the given text. The decoder module takes the concatenated features from the speaker and text encoders and generates the mel spectrogram corresponding to the prescribed text. Finally, the vocoder transforms the mel spectrogram into audible audio. In this process, pretrained speaker encoders and vocoders are employed \cite{wang2017tacotron}. The training process primarily focuses on optimizing the text encoder and decoder components. The second direction of TTS is referred to as end-to-end models, which eliminate the need for generating mel-spectrograms in the intermediate process \cite{kim2021conditional}. Instead, these models directly generate waveforms.

The process of audio-driven talking head generation encompasses two primary methodologies: image-based methods and video-based methods. The objective of audio-driven talking head generation is to attain a seamless integration between the input audio and the generated visual content. Through the utilization of either the image-based or video-based approach, researchers have made remarkable progress in generating persuasive and synchronized talking head videos.

The image based method includes two sub-categories: the 2D method and the 3D method, as illustrated in part (c) of  Figure \ref{fig 1: Comparsion of previous and current TTV and its components}. In the 2D method, also known as the flow field method, two encoders are employed to extract features from the input image and audio, respectively. These extracted features are then utilized to predict the head motion sequence. By generating the head motion sequence, the flow field can be predicted, which is subsequently applied in the video generation process. The flow field serves as a crucial component for synthesizing realistic and synchronized movements of the talking head. This approach leverages the interplay between image and audio information to generate dynamic visual content. On the other hand, the 3D method, referred to as the 3D Morphable Model (3DMM) method, relies on a different approach. Initially, 3DMM coefficients are extracted from the input image, which represent the geometric and appearance characteristics of the head. By combining these derived 3DMM coefficients with the audio input, sequences of 3DMM coefficients can be obtained. Using 3DMM and rendering neural networks, these sequences are then used to generate the final video output. The 3DMM method takes advantage of the underlying 3D structure of the head and its variations to synthesize realistic visual representations.

The video-based method for audio-driven talking head generation involves a series of three sequential tasks aimed at producing high-quality videos. These tasks include: face video generation with a canonical expression, audio-driven lip-sync, and face enhancement for enhancing photo-realism. To initiate the process, an expression editing network is employed to generate a video with a predefined canonical expression. In the next task, the edited video, along with the provided audio, is inputted into a lip-sync network. This network aligns the acoustic features of the audio with the visual cues of the video, resulting in a lip-synced video where the facial movements accurately synchronize with the spoken words. Lastly, an identity-aware face enhancement network is employed to refine the quality and realism of the generated video.

\section{Experiments}

In this section, we will provide the details of the experiments conducted. For audio synthesis, we selected the first 90 seconds of Steve Jobs' speech at Stanford in 2005, which can be found in the following link: \href{https://www.youtube.com/watch?v=UF8uR6Z6KLc}{watch video}. The text chosen for speech synthesis is a concise introduction to deep learning, which is as follows: \textit{The adjective deep in deep learning refers to the use of multiple layers in the network. Early work showed that a linear perceptron cannot be a universal classifier, but that a network with a nonpolynomial activation function with one hidden layer of unbounded width can. Deep learning is a modern variation that is concerned with an unbounded number of layers of bounded size, which permits practical application and optimized implementation, while retaining theoretical universality under mild conditions}.

For the audio-driven talking head generation, we utilized the image displayed in part (a) of Figure \ref{fig 3: Comparsion of different audio-driven talking head generation}. Specifically, two images of Steve Jobs were chosen for this experiment. Furthermore, a video featuring Jobs in 1996 as shown in the following link: \href{https://www.youtube.com/watch?v=SgWdjvRgouk}{watch video} has been selected as the primary video reference, with his face occupying the majority of the footage, sacrificing other elements. To maintain consistency, the same audio source has been chosen for the synthesis of new videos by combining it with the previously mentioned video. 

In our TTS experiments, we employed four different methods: Tacotron, Vits, Tortoise, and YourTTS. Each of these methods contributes unique features to the audio generation process. To ensure consistency, all generated audios were resampled to have a total length of 70,000 frames as shown in \ref{fig 2: Comparsion of different TTS for generating audios}. Tacotron, Vits, and Tortoise successfully achieve the goal of TTS by generating high-quality audios. However, they lack the ability to specify the identity of the speakers. Despite their impressive audio quality, the voices produced by these methods do not resemble that of Steve Jobs.

On the other hand, YourTTS is designed to learn and mimic the distinct voice of Steve Jobs. As a result, the generated audios exhibit a closer resemblance to Steve Jobs' voice. However, it is important to note that the audio quality may slightly deteriorate in the process. In our current work, our focus lies primarily on the qualitative comparison of these different methods. We aim to assess the similarities and differences in the generated audios, particularly in terms of their resemblance to Steve Jobs' voice. However, in our future work, we plan to conduct a more comprehensive quantitative analysis, evaluating the performance of these methods using objective metrics. Additionally, we will explore the inclusion of more audio-agnostic TTS methods, further expanding the scope of our research.

\begin{figure}[tb!]
\centering
    \includegraphics[width=0.99\columnwidth]{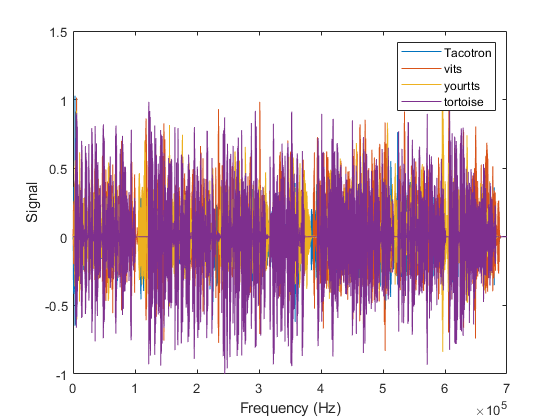}
\end{figure}

This section presents a comparative analysis of three different methods for audio-driven talking head generation: Audio2Head, StyleHEAT, and SadTalker. The results of these methods are shown in Figure \ref{fig 3: Comparsion of different audio-driven talking head generation}.

Audio2Head demonstrates the ability to generate images with lip synchronization, resulting in a reasonable match between the audio and the generated visuals. However, the image quality of Audio2Head falls short, exhibiting blurriness and a lack of sharpness. Moreover, the generated videos exhibit substantial motion changes, with discontinuous shoulder movements across different frames.

StyleHEAT, unfortunately, faces more severe issues. The method tends to produce distorted and visually unpleasant images. It appears that StyleHEAT lacks a face landmark identification module, leading to the generation of multiple sets of eyes within a single image.

In contrast, SadTalker showcases high-quality image generation and successful lip synchronization with the audio. However, a noteworthy limitation of SadTalker is that it produces relatively minimal movements of "Steve Jobs" in the generated videos. This poses a challenge when attempting to incorporate significant movements and gestures for an accurate portrayal.

The evaluation and comparison of these methods shed light on their respective strengths and weaknesses. It provides valuable insights into the trade-offs between image quality, lip synchronization, and the fidelity of motion representation. 

Another work, i.e., VideoRetalking is conducted on video generation based on previous videos. In our assessment, this method exhibited improvements compared to previous works, as it effectively incorporates a broader range of valuable information in the generated videos, encompassing enhanced gestures and emotions. However, challenges arose when applying this method to videos featuring Jobs due to his significant movements in the footage and the low resolution of the existing videos, resulting in a degradation of the generated video's consistency. Subsequent tests conducted on modern, high-quality videos with relatively smaller movements yielded more favorable outcomes. Moving forward, future research will concentrate on addressing these challenges and enhancing the overall quality of the generated videos.

\begin{figure}[tb!]
    \centering
    \includegraphics[width=0.99\columnwidth]{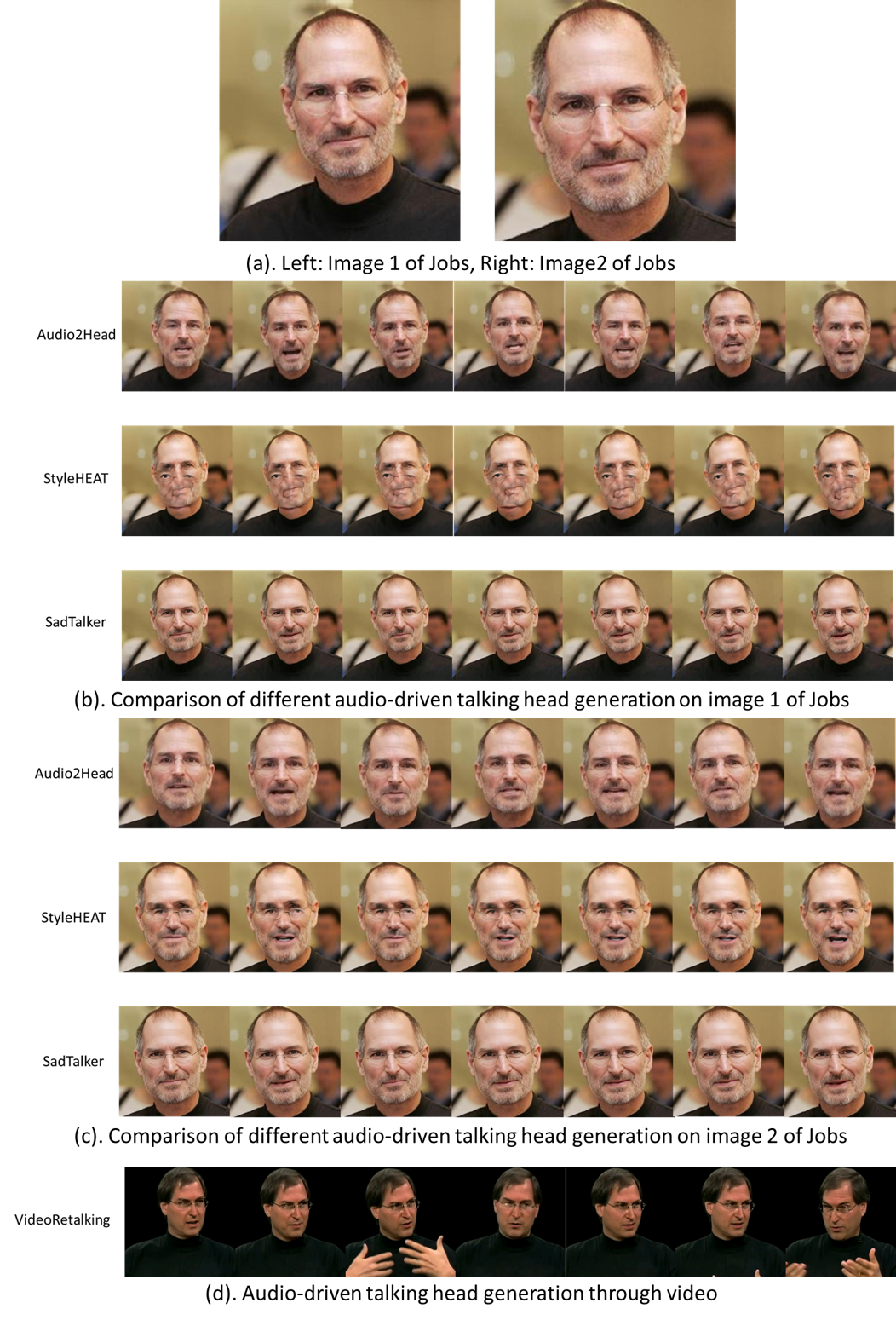}
	\caption{Comparsion of different audio-driven talking head generation}
	\label{fig 3: Comparsion of different audio-driven talking head generation}
\end{figure}

\section{Future Work}

In addition to the aforementioned approaches, several other methods have been explored for TTS and audio-driven talking head generation. These alternative techniques offer unique advantages and pave the way for further advancements in the field.

One notable method in TTS is VALL-E, which introduces a novel approach that eliminates the need for transforming audio into the spectrogram domain. Instead, it leverages a neural codec as an intermediate representation, enabling more efficient and accurate synthesis. Notably, VALL-E is trained on a significantly larger dataset, contributing to enhanced performance and naturalness in speech generation.

When exploring the realm of audio-driven talking head generation, several other methods have also been developed that utilize images or videos as input to generate videos. To gain a deeper understanding and draw meaningful conclusions, a comprehensive comparison of these methods should be conducted. This comparison would allow for an in-depth analysis of the strengths and weaknesses of each approach, enabling researchers to identify the most effective and suitable techniques for different applications within the field of audio-driven talking head generation.

In our future work, we intend to explore and evaluate these alternative methods, including VALL-E and video-based approaches, to further enhance the quality and naturalness of TTV synthesis. By testing and comparing these methods, we aim to uncover novel techniques that can push the boundaries of TTV and deliver high-quality results in terms of visual and auditory coherence.

\section{Conclusion}
This paper introduces a pioneering approach in zero-shot identity-agonistic TTV generation. The proposed method enables the generation of high-quality videos by leveraging text input, along with a pre-existing audio clip and image of the desired identity. The obtained results demonstrate the capability to produce visually impressive videos when all the necessary information is provided.

Further research will be conducted to enhance the performance of the proposed TTV framework by exploring the most effective techniques in TTS and audio-driven talking head generation. To provide a comprehensive evaluation of the generated videos, additional quantitative metrics will be incorporated.


\bibliographystyle{ACM-Reference-Format}
\bibliography{main}


\begin{thebibliography}{18}


\ifx \showCODEN    \undefined \def \showCODEN     #1{\unskip}     \fi
\ifx \showDOI      \undefined \def \showDOI       #1{#1}\fi
\ifx \showISBNx    \undefined \def \showISBNx     #1{\unskip}     \fi
\ifx \showISBNxiii \undefined \def \showISBNxiii  #1{\unskip}     \fi
\ifx \showISSN     \undefined \def \showISSN      #1{\unskip}     \fi
\ifx \showLCCN     \undefined \def \showLCCN      #1{\unskip}     \fi
\ifx \shownote     \undefined \def \shownote      #1{#1}          \fi
\ifx \showarticletitle \undefined \def \showarticletitle #1{#1}   \fi
\ifx \showURL      \undefined \def \showURL       {\relax}        \fi
\providecommand\bibfield[2]{#2}
\providecommand\bibinfo[2]{#2}
\providecommand\natexlab[1]{#1}
\providecommand\showeprint[2][]{arXiv:#2}

\bibitem[Casanova et~al\mbox{.}(2023)]%
        {casanova2023yourTTS}
\bibfield{author}{\bibinfo{person}{Edresson Casanova}, \bibinfo{person}{Julian
  Weber}, \bibinfo{person}{Christopher Shulby},
  \bibinfo{person}{Arnaldo~Candido Junior}, \bibinfo{person}{Eren Gölge},
  {and} \bibinfo{person}{Moacir~Antonelli Ponti}.}
  \bibinfo{year}{2023}\natexlab{}.
\newblock \bibinfo{title}{YourTTS: Towards Zero-Shot Multi-Speaker TTS and
  Zero-Shot Voice Conversion for everyone}.
\newblock
\newblock
\showeprint[arxiv]{2112.02418}~[cs.SD]


\bibitem[Cheng et~al\mbox{.}(2022)]%
        {cheng2022videoretalking}
\bibfield{author}{\bibinfo{person}{Kun Cheng}, \bibinfo{person}{Xiaodong Cun},
  \bibinfo{person}{Yong Zhang}, \bibinfo{person}{Menghan Xia},
  \bibinfo{person}{Fei Yin}, \bibinfo{person}{Mingrui Zhu},
  \bibinfo{person}{Xuan Wang}, \bibinfo{person}{Jue Wang}, {and}
  \bibinfo{person}{Nannan Wang}.} \bibinfo{year}{2022}\natexlab{}.
\newblock \bibinfo{title}{VideoReTalking: Audio-based Lip Synchronization for
  Talking Head Video Editing In the Wild}.
\newblock
\newblock
\showeprint[arxiv]{2211.14758}~[cs.CV]


\bibitem[Guo et~al\mbox{.}(2021)]%
        {guo2021adnerf}
\bibfield{author}{\bibinfo{person}{Yudong Guo}, \bibinfo{person}{Keyu Chen},
  \bibinfo{person}{Sen Liang}, \bibinfo{person}{Yong-Jin Liu},
  \bibinfo{person}{Hujun Bao}, {and} \bibinfo{person}{Juyong Zhang}.}
  \bibinfo{year}{2021}\natexlab{}.
\newblock \bibinfo{title}{AD-NeRF: Audio Driven Neural Radiance Fields for
  Talking Head Synthesis}.
\newblock
\newblock
\showeprint[arxiv]{2103.11078}~[cs.CV]


\bibitem[Heigold et~al\mbox{.}(2016)]%
        {heigold2016end}
\bibfield{author}{\bibinfo{person}{Georg Heigold}, \bibinfo{person}{Ignacio
  Moreno}, \bibinfo{person}{Samy Bengio}, {and} \bibinfo{person}{Noam
  Shazeer}.} \bibinfo{year}{2016}\natexlab{}.
\newblock \showarticletitle{End-to-end text-dependent speaker verification}. In
  \bibinfo{booktitle}{\emph{2016 IEEE International Conference on Acoustics,
  Speech and Signal Processing (ICASSP)}}. IEEE, \bibinfo{pages}{5115--5119}.
\newblock


\bibitem[Kim et~al\mbox{.}(2021)]%
        {kim2021conditional}
\bibfield{author}{\bibinfo{person}{Jaehyeon Kim}, \bibinfo{person}{Jungil
  Kong}, {and} \bibinfo{person}{Juhee Son}.} \bibinfo{year}{2021}\natexlab{}.
\newblock \bibinfo{title}{Conditional Variational Autoencoder with Adversarial
  Learning for End-to-End Text-to-Speech}.
\newblock
\newblock
\showeprint[arxiv]{2106.06103}~[cs.SD]


\bibitem[Kumar et~al\mbox{.}(2017)]%
        {kumar2017obamanet}
\bibfield{author}{\bibinfo{person}{Rithesh Kumar}, \bibinfo{person}{Jose
  Sotelo}, \bibinfo{person}{Kundan Kumar}, \bibinfo{person}{Alexandre de
  Brebisson}, {and} \bibinfo{person}{Yoshua Bengio}.}
  \bibinfo{year}{2017}\natexlab{}.
\newblock \bibinfo{title}{ObamaNet: Photo-realistic lip-sync from text}.
\newblock
\newblock
\showeprint[arxiv]{1801.01442}~[cs.CV]


\bibitem[Li et~al\mbox{.}(2021)]%
        {li2021writeaspeaker}
\bibfield{author}{\bibinfo{person}{Lincheng Li}, \bibinfo{person}{Suzhen Wang},
  \bibinfo{person}{Zhimeng Zhang}, \bibinfo{person}{Yu Ding},
  \bibinfo{person}{Yixing Zheng}, \bibinfo{person}{Xin Yu}, {and}
  \bibinfo{person}{Changjie Fan}.} \bibinfo{year}{2021}\natexlab{}.
\newblock \bibinfo{title}{Write-a-speaker: Text-based Emotional and Rhythmic
  Talking-head Generation}.
\newblock
\newblock
\showeprint[arxiv]{2104.07995}~[cs.CV]


\bibitem[Oord et~al\mbox{.}(2016)]%
        {oord2016wavenet}
\bibfield{author}{\bibinfo{person}{Aaron van~den Oord}, \bibinfo{person}{Sander
  Dieleman}, \bibinfo{person}{Heiga Zen}, \bibinfo{person}{Karen Simonyan},
  \bibinfo{person}{Oriol Vinyals}, \bibinfo{person}{Alex Graves},
  \bibinfo{person}{Nal Kalchbrenner}, \bibinfo{person}{Andrew Senior}, {and}
  \bibinfo{person}{Koray Kavukcuoglu}.} \bibinfo{year}{2016}\natexlab{}.
\newblock \showarticletitle{Wavenet: A generative model for raw audio}.
\newblock \bibinfo{journal}{\emph{arXiv preprint arXiv:1609.03499}}
  (\bibinfo{year}{2016}).
\newblock


\bibitem[Ren et~al\mbox{.}(2019)]%
        {ren2019fastspeech}
\bibfield{author}{\bibinfo{person}{Yi Ren}, \bibinfo{person}{Yangjun Ruan},
  \bibinfo{person}{Xu Tan}, \bibinfo{person}{Tao Qin}, \bibinfo{person}{Sheng
  Zhao}, \bibinfo{person}{Zhou Zhao}, {and} \bibinfo{person}{Tie-Yan Liu}.}
  \bibinfo{year}{2019}\natexlab{}.
\newblock \bibinfo{title}{FastSpeech: Fast, Robust and Controllable Text to
  Speech}.
\newblock
\newblock
\showeprint[arxiv]{1905.09263}~[cs.CL]


\bibitem[Tang et~al\mbox{.}(2022a)]%
        {tang2022memories}
\bibfield{author}{\bibinfo{person}{Anni Tang}, \bibinfo{person}{Tianyu He},
  \bibinfo{person}{Xu Tan}, \bibinfo{person}{Jun Ling}, \bibinfo{person}{Runnan
  Li}, \bibinfo{person}{Sheng Zhao}, \bibinfo{person}{Li Song}, {and}
  \bibinfo{person}{Jiang Bian}.} \bibinfo{year}{2022}\natexlab{a}.
\newblock \bibinfo{title}{Memories are One-to-Many Mapping Alleviators in
  Talking Face Generation}.
\newblock
\newblock
\showeprint[arxiv]{2212.05005}~[cs.CV]


\bibitem[Tang et~al\mbox{.}(2022b)]%
        {tang2022realtime}
\bibfield{author}{\bibinfo{person}{Jiaxiang Tang}, \bibinfo{person}{Kaisiyuan
  Wang}, \bibinfo{person}{Hang Zhou}, \bibinfo{person}{Xiaokang Chen},
  \bibinfo{person}{Dongliang He}, \bibinfo{person}{Tianshu Hu},
  \bibinfo{person}{Jingtuo Liu}, \bibinfo{person}{Gang Zeng}, {and}
  \bibinfo{person}{Jingdong Wang}.} \bibinfo{year}{2022}\natexlab{b}.
\newblock \bibinfo{title}{Real-time Neural Radiance Talking Portrait Synthesis
  via Audio-spatial Decomposition}.
\newblock
\newblock
\showeprint[arxiv]{2211.12368}~[cs.CV]


\bibitem[Wang et~al\mbox{.}(2023)]%
        {wang2023neural}
\bibfield{author}{\bibinfo{person}{Chengyi Wang}, \bibinfo{person}{Sanyuan
  Chen}, \bibinfo{person}{Yu Wu}, \bibinfo{person}{Ziqiang Zhang},
  \bibinfo{person}{Long Zhou}, \bibinfo{person}{Shujie Liu},
  \bibinfo{person}{Zhuo Chen}, \bibinfo{person}{Yanqing Liu},
  \bibinfo{person}{Huaming Wang}, \bibinfo{person}{Jinyu Li},
  \bibinfo{person}{Lei He}, \bibinfo{person}{Sheng Zhao}, {and}
  \bibinfo{person}{Furu Wei}.} \bibinfo{year}{2023}\natexlab{}.
\newblock \bibinfo{title}{Neural Codec Language Models are Zero-Shot Text to
  Speech Synthesizers}.
\newblock
\newblock
\showeprint[arxiv]{2301.02111}~[cs.CL]


\bibitem[Wang et~al\mbox{.}(2021b)]%
        {wang2021audio2head}
\bibfield{author}{\bibinfo{person}{Suzhen Wang}, \bibinfo{person}{Lincheng Li},
  \bibinfo{person}{Yu Ding}, \bibinfo{person}{Changjie Fan}, {and}
  \bibinfo{person}{Xin Yu}.} \bibinfo{year}{2021}\natexlab{b}.
\newblock \bibinfo{title}{Audio2Head: Audio-driven One-shot Talking-head
  Generation with Natural Head Motion}.
\newblock
\newblock
\showeprint[arxiv]{2107.09293}~[cs.CV]


\bibitem[Wang et~al\mbox{.}(2021a)]%
        {wang2021oneshot}
\bibfield{author}{\bibinfo{person}{Suzhen Wang}, \bibinfo{person}{Lincheng Li},
  \bibinfo{person}{Yu Ding}, {and} \bibinfo{person}{Xin Yu}.}
  \bibinfo{year}{2021}\natexlab{a}.
\newblock \bibinfo{title}{One-shot Talking Face Generation from Single-speaker
  Audio-Visual Correlation Learning}.
\newblock
\newblock
\showeprint[arxiv]{2112.02749}~[cs.CV]


\bibitem[Wang et~al\mbox{.}(2017)]%
        {wang2017tacotron}
\bibfield{author}{\bibinfo{person}{Yuxuan Wang}, \bibinfo{person}{RJ
  Skerry-Ryan}, \bibinfo{person}{Daisy Stanton}, \bibinfo{person}{Yonghui Wu},
  \bibinfo{person}{Ron~J. Weiss}, \bibinfo{person}{Navdeep Jaitly},
  \bibinfo{person}{Zongheng Yang}, \bibinfo{person}{Ying Xiao},
  \bibinfo{person}{Zhifeng Chen}, \bibinfo{person}{Samy Bengio},
  \bibinfo{person}{Quoc Le}, \bibinfo{person}{Yannis Agiomyrgiannakis},
  \bibinfo{person}{Rob Clark}, {and} \bibinfo{person}{Rif~A. Saurous}.}
  \bibinfo{year}{2017}\natexlab{}.
\newblock \bibinfo{title}{Tacotron: Towards End-to-End Speech Synthesis}.
\newblock
\newblock
\showeprint[arxiv]{1703.10135}~[cs.CL]


\bibitem[Yin et~al\mbox{.}(2022)]%
        {yin2022styleheat}
\bibfield{author}{\bibinfo{person}{Fei Yin}, \bibinfo{person}{Yong Zhang},
  \bibinfo{person}{Xiaodong Cun}, \bibinfo{person}{Mingdeng Cao},
  \bibinfo{person}{Yanbo Fan}, \bibinfo{person}{Xuan Wang},
  \bibinfo{person}{Qingyan Bai}, \bibinfo{person}{Baoyuan Wu},
  \bibinfo{person}{Jue Wang}, {and} \bibinfo{person}{Yujiu Yang}.}
  \bibinfo{year}{2022}\natexlab{}.
\newblock \bibinfo{title}{StyleHEAT: One-Shot High-Resolution Editable Talking
  Face Generation via Pre-trained StyleGAN}.
\newblock
\newblock
\showeprint[arxiv]{2203.04036}~[cs.CV]


\bibitem[Zhang et~al\mbox{.}(2022)]%
        {zhang2022text2video}
\bibfield{author}{\bibinfo{person}{Sibo Zhang}, \bibinfo{person}{Jiahong Yuan},
  \bibinfo{person}{Miao Liao}, {and} \bibinfo{person}{Liangjun Zhang}.}
  \bibinfo{year}{2022}\natexlab{}.
\newblock \bibinfo{title}{Text2Video: Text-driven Talking-head Video Synthesis
  with Personalized Phoneme-Pose Dictionary}.
\newblock
\newblock
\showeprint[arxiv]{2104.14631}~[cs.CV]


\bibitem[Zhang et~al\mbox{.}(2023)]%
        {zhang2023sadtalker}
\bibfield{author}{\bibinfo{person}{Wenxuan Zhang}, \bibinfo{person}{Xiaodong
  Cun}, \bibinfo{person}{Xuan Wang}, \bibinfo{person}{Yong Zhang},
  \bibinfo{person}{Xi Shen}, \bibinfo{person}{Yu Guo}, \bibinfo{person}{Ying
  Shan}, {and} \bibinfo{person}{Fei Wang}.} \bibinfo{year}{2023}\natexlab{}.
\newblock \bibinfo{title}{SadTalker: Learning Realistic 3D Motion Coefficients
  for Stylized Audio-Driven Single Image Talking Face Animation}.
\newblock
\newblock
\showeprint[arxiv]{2211.12194}~[cs.CV]


\end{thebibliography}

\end{document}